%% file: main.tex
\pdfoutput=1

\documentclass[11pt]{article}

\def\@fnsymbol#1{\ensuremath{\ifcase#1\or \dagger\or \ddagger\or
   \mathsection\or \mathparagraph\or \|\or **\or \dagger\dagger
   \or \ddagger\ddagger \else\@ctrerr\fi}}

\usepackage[]{EMNLP2022}

\usepackage{subcaption}

\usepackage{times}
\usepackage{latexsym}

\usepackage{soul}

\usepackage[T1]{fontenc}

\usepackage[utf8]{inputenc}

\usepackage{microtype}

\usepackage{inconsolata}

\input{tex/commands}

\title{System Demo: Tool and Infrastructure for \\ Offensive Language Error Analysis (\lib{}) in English}

\author{Marie Grace\thanks{\indent These authors contributed equally.} , Xajavion ``Jay'' Seabrum\footnotemark[1], Dananjay Srinivas\footnotemark[1], Alexis Palmer \\
University of Colorado Boulder \\
\texttt{first.last@colorado.edu};  
 \texttt{olea.ask@gmail.com}\thanks{\indent Please direct inquiries about the library to this email.}}

\begin{document}
\maketitle
\input{tex/abstract}
\input{tex/intro}
\input{tex/related_work}
\input{tex/system}

\input{tex/use_cases}

\input{tex/discussion_conclusion}
\input{tex/endmatter}

\bibliography{anthology,custom}
\bibliographystyle{acl_natbib}

\input{tex/appendix}

\end{document}

%% file: tex/commands.tex
\usepackage{xcolor,colortbl}
\usepackage{todonotes}
\usepackage{multicol}

\newcommand{\lib}[0]{OLEA}

\definecolor{alexis}{rgb}{0.06, 0.75, 0.99}
\definecolor{jay}{rgb}{0.85, 0.65, 0.13} 
\definecolor{marie}{rgb}{0.65, 0, 0.35} 
\definecolor{ds}{rgb}{255,0,0}
\definecolor{mdgreen}{rgb}{0,0.6,0}
\definecolor{orange}{rgb}{1,0.5,0}

\newcounter{example}

\newenvironment{packed_enum}{
\begin{enumerate}
  \setlength{\itemsep}{1pt}
  \setlength{\parskip}{0pt}
  \setlength{\parsep}{0pt}
  \setlength{\leftmargin}{0pt}
}{\end{enumerate}}

\definecolor{codegreen}{rgb}{0,0.6,0}
\definecolor{codegray}{rgb}{0.5,0.5,0.5}
\definecolor{codepurple}{rgb}{0.58,0,0.82}
\definecolor{backcolour}{rgb}{0.95,0.95,0.92}
\usepackage{listings}
\usepackage{xcolor}

\lstdefinestyle{mystyle}{
    backgroundcolor=\color{backcolour},   
    commentstyle=\color{codegreen},
    keywordstyle=\color{magenta},
    numberstyle=\tiny\color{codegray},
    stringstyle=\color{codepurple},
    basicstyle=\ttfamily\footnotesize,
    breakatwhitespace=false,         
    breaklines=true,                 
    captionpos=b,                    
    keepspaces=true,                 
    numbers=left,                    
    numbersep=5pt,                  
    showspaces=false,                
    showstringspaces=false,
    showtabs=false,                  
    tabsize=2
}

%% file: tex/abstract.tex
\begin{abstract}
The automatic detection of offensive language is a pressing societal need. 
Many systems perform well on explicit offensive language but struggle to detect more complex, nuanced, or implicit cases of offensive and hateful language.
\lib{} is an open-source Python library that provides easy-to-use tools for error analysis in the context of detecting offensive language in English.
\lib{} also provides an infrastructure for re-distribution of new datasets and analysis methods requiring very little coding.
\end{abstract}

%% file: tex/intro.tex
\section{Introduction}\label{sec:intro}

Offensive language\footnote{We use the term "offensive language" to encompass offensive language and hate speech. This paper contains censored offensive language examples.} detection models have become integral to online platforms' moderation systems.  
Such systems excel at detecting and filtering out messages with explicit keywords and mentions; however, these systems are known (1) to perform poorly on messages that are implicitly offensive and that have negation \cite{rottger2020hatecheck,palmer2020cold}; (2) to have annotator biases affecting the detection of offensive language \cite{sap2021annotators}; (3) not to be robust to diachronic language and its usage \cite{florio2020time}; and (4) to be insensitive to and overdetect AAE as offensive language \cite{sap2021annotators, blodgett-etal-2016-demographic}.  
These issues and gaps are important to recognize, as failing to address them can cause marginalized groups to be further dehumanized or attacked \cite{mathew2021hatexplain, kennedy2020constructing}. 

All of these complexities underscore the importance of having a detailed and nuanced understanding of the true capabilities of models for automatic detection of offensive language. 
We need to move away from viewing the task primarily as a binary classification problem \cite{kennedy2020constructing}, and we need to consider more than a single F1 score when evaluating our models.
We also need a cohesive and common framework for error analysis, a tool to help researchers more easily understand why their systems fail to detect certain types of offensive speech \cite{poletto2021resources}.
We need a better understanding of model performance on out-of-domain data if we are to achieve robust models that are quickly deployable on new domains.
Finally, we need to address some of the practical hurdles that prevent more widespread adoption of some datasets, taking advantage of recent improvements in NLP research infrastructure.


With these issues in mind, we introduce \lib{}\footnote{ \url{https://pypi.org/project/olea/}}\textsuperscript{,}\footnote{ \url{https://www.youtube.com/watch?v=730MZktD5q4}}, an open-source Offensive Language Error Analysis tool and infrastructure, providing:
\begin{packed_enum}
\item easy-to-use methods for error analysis and evaluation of new models on existing diagnostic datasets, including model comparison;
\item interfaces to two diagnostic datasets focused on nuanced linguistic analysis; and
\item scaffolding to support easy distribution of new datasets and associated analysis methods.
\end{packed_enum}

\noindent
The error analysis and evaluation tools provide insights into where and how models can be improved.
The infrastructure provided by \lib{} helps researchers distribute datasets to the broader research community, together with dataset-specific analysis methods, all with very low overhead.

%% file: tex/related_work.tex
\section{Background and Related Work}\label{sec:relwork}

\input{imgs/SystemFlow}

Offensive language is complex, and systems for detecting it automatically need to be able to handle both explicit \textit{and}
implicit cases \cite{schmidt-wiegand-2017-survey, waseem-etal-2017-understanding}.  
Detecting explicit offensive language often relies on keyword detection \cite{wiegand-etal-2019-detection}, but keyword-driven systems can lead to messages being falsely flagged, causing unchecked or unnoticed racial biases to propagate in the system's decisions \cite{sap2021annotators, blodgett-etal-2016-demographic}. 
Implicit offensive language is generally more difficult to detect
than its explicit counterpart \cite{elsherief-etal-2021-latent, caselli-etal-2020-feel}.  
Furthermore, it is more likely to change over time given real world circumstances and coinages of new phrases and terms to implicitly refer to minority groups \cite{florio2020time}.  

Datasets for this task
take varied approaches.
For example, HateXplain \cite{mathew2021hatexplain} and CAD \cite{vidgen-etal-2021-introducing} both provide rationales indicating where annotators see offensive content.
OLID \cite{zampierietal2019} 
identifies offensive text and the specific targeted minority group in a three-tiered labeling structure.
HateCheck \cite{rottger2020hatecheck} and COLD \cite{palmer2020cold} are described more in Section~\ref{sec:data}.
Because many of these datasets address different (often overlapping) concerns,
direct comparison is difficult.

Additionally, linguistic explainability of the prediction \textit{failures} of NLP models has lagged behind their performance as measured against benchmark datasets \cite{hovy-2022}.
To address part of this concern, \citet{mcmillan-major-etal-2022-interactive} provide an interactive system mostly for end users of offensive language detection systems. 
Their system helps users explore datasets and understand how individual text inputs are scored and classified by different models.
\lib{} has complementary functionality, focusing on fine-grained analysis of model performance (especially misclassifications) across existing evaluation datasets.
We focus on model developers rather than end users, providing streamlined error analysis and interpretation of system outputs relative to linguistically-grounded categorizations. 

%% file: imgs/SystemFlow.tex
    \begin{figure*}[t!]\centering
    \includegraphics[width = 0.85\textwidth, height = 0.12\textheight]{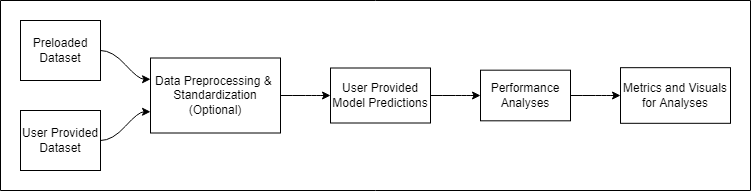}
    \caption{Diagram of the \lib{} Library pipeline.}
    \label{fig:systemflow}
    \end{figure*}

%% file: tex/system.tex
\section{Library Tour and Design}\label{sec:library}
 

Figure \ref{fig:systemflow} shows an overview of \lib{}'s core functionalities.
Users submit their model's predictions (\ref{sec:prelim}) on \lib{}'s preloaded datasets (\ref{sec:data}) and then call error analysis and evaluation functions (\ref{sec:EA}).
Users may also extend \lib{} with new datasets and may write new analysis functions, adding to the library's capabilities (\ref{sec:extend}).

\lib{} can be installed using pip (\texttt{pip install olea}).
Most of \lib{}'s modules expect a Pandas\footnote{\url{https://pandas.pydata.org/docs/}} dataframe 
with the text of the instance to be classified, one or more labels indicating offensiveness, and a predicted label for the instance.
Dataframes may include columns with additional information related to the instance and/or its annotation.

\subsection{Preloaded diagnostic datasets}\label{sec:data}

The primary function of \lib{} is to make it easy for users to evaluate the capabilities of their models in a fine-grained way.
We provide interfaces (via HuggingFace's datasets library\footnote{\url{https://huggingface.co/docs/datasets/index}} and the HuggingFace Hub) to two datasets designed specifically for diagnostic evaluation of offensive language detection systems.
Both datasets include fine-grained annotations and binary offensiveness labels and were curated to compare model performance with linguistic phenomena. 
Appendices~\ref{app:coldDim} and \ref{app:hcDim} list the features available for analysis.

\textbf{HateCheck} \cite{rottger2020hatecheck} provides a test suite that includes labels reflecting some of the specific linguistic constructions often seen in online hate speech, such as use of spelling changes to obscure hateful language and presence of threatening language. 
HateCheck also includes annotations of the specific identities targeted in each instance of hate speech. 
\textbf{COLD} \cite{palmer2020cold} provides fine-grained labels, from multiple annotators, of some linguistic phenomena relevant for implicit hate speech and offensive language.
Some examples are presence/absence of slur terms and presence/absence of adjectival nominalizations.


\subsection{Submitting predictions}\label{sec:prelim}

Before using the analysis functions described below, the user needs to submit their model's predictions on the selected dataset, as well as a mapping between the model's predicted labels (e.g. 1, 0) and the labels in the selected dataset (e.g. \texttt{hateful}, \texttt{non-hateful}).\footnote{Note that this process applies both for preloaded datasets and for datasets read in from the user's own system.}
The code snippet below illustrates the process and assumes that the user's data has been stored as a Pandas dataframe named \texttt{user\_data}. 
In this example, the model predictions are found in a column called \texttt{predictions}.
The user has selected three features for potential analysis: \texttt{Text}, \texttt{is\_slur} and \texttt{text\_length}.

\input{code_examples/entry}

\noindent
The \texttt{submit} method passes the relevant parameters to the analysis module. 

\subsection{Error analysis functions}\label{sec:EA}

The heart of our library is a collection of functions for detailed evaluation and error analysis.
\textit{Throughout, we evaluate the model's coarse-grained classification performance (e.g. offensive vs. not offensive) for subsets of instances grouped according to a particular feature.
The features generally correspond to dataframe columns.
For example, we may compare performance for instances containing a slur term to performance for instances with no slur term.}
Plots are produced using Matplotlib \cite{matplotlib-Hunter:2007}, and we include the option to save plots to files.
Section~\ref{sec:usecases} shows concrete examples of the analysis outputs, and code examples appear in Appendix~\ref{sec:appendix}.

\paragraph{\texttt{analyze\_on}.}
In its most general version, this function evaluates model performance for a categorical column specified by the user.
\lib{} includes versions of \texttt{analyze\_on} customized to the two preloaded datasets.
The COLD-specific version evaluates performance for features constructed from combinations of four binary features: offensiveness, presence of slur term, presence of adjectival nominalization, and presence of linguistic distancing.
The HateCheck-specific version includes linguistic features (e.g. negation, derogation, or profanity) and features related to the identity of the targeted individual or class (e.g. trans people, Muslims, or disabled people).

\paragraph{\texttt{check\_anno\_agreement}.}
This function is intended for datasets which include labels from multiple annotators, such as COLD. 
The function compares performance on instances with full annotator agreement for the label of offensiveness to performance on instances with partial agreement.
Full annotator agreement is taken as a proxy for instances that are ``easy'' to classify, and partial agreement indicates more complex cases.

\paragraph{\texttt{aave}.}
This function evaluates performance for instances (likely) written using African American English.
The scores are calculated using the TwitterAAE model \cite{blodgett-etal-2016-demographic}. 
These scores represent an inference of the proportion of words in the instance that come from a demographically-associated language/dialect.

\paragraph{\texttt{check\_substring}.}
Given a user-specified text string, this function compares performance on instances with the substring to instances without.

\paragraph{\texttt{str\_len\_analysis}.}
This function outputs a histogram showing model performance on instances of different lengths (character or word count).

\subsection{Adding datasets and analyses}\label{sec:extend}

Extensibility is a key principle guiding the design of \lib{}, with the goal of providing an easy-to-use, uniform platform for error analysis in the context of offensive language detection.
In addition to the two preloaded datasets, users can submit their own datasets using the process described in \ref{sec:ownData}.
Modifying a single Python class enables a built-in suite of analysis functions.

\lib{} has a helper function for preprocessing English text to remove user names and URLs and convert emoji to their textual descriptions.\footnote{Preprocessing scheme is described in \citet{palmer2020cold}.}

\input{code_examples/preprocess_text}


\noindent
For example, the text preprocessor will convert "@username\_1 Have you seen the video that @another\_user made?  :eyes: :fire: :hundred-points: https://fakelink.io" to "USER have you seen the video that USER made? eyes fire hundred points HTML".

Finally, users can write and share their own analysis functions, focusing on user-specified dimensions, as in \ref{sec:sharing}. 
\lib{}'s code is modularized such that adding a new analysis requires enough Python knowledge to write the function, but not a detailed understanding of the entire codebase.





%% file: code_examples/entry.tex
\begin{lstlisting}[language=Python]
from olea.data import Dataset
setup = Dataset(
          data = user_data,
          features = ["Text","is_slur", 
                     "text_length"],
          gold_column = "gold_labels", 
          text_column = "Text")
predictions = user_data["predictions"]
mapping = {"hateful": 1,
           "non-hateful": 0}
data_submit = setup.submit(
            batch = user_data, 
            predictions = predictions,
            map = mapping)
\end{lstlisting}

%% file: code_examples/preprocess_text.tex
\begin{lstlisting}[language=Python]
from olea.utils.preprocess_text import PreprocessText as pt
processed_text = pt.execute(user_data["raw_text"])
user_data["preprocessed_text"] = preprocessed_text
\end{lstlisting}

%% file: tex/use_cases.tex
\section{Use Case Demonstrations}\label{sec:usecases}
The three main use cases envisioned are: a) analysis on preloaded datasets (\ref{subsec:generic_analysis}), including model comparison (\ref{sec:compare}), b) analysis on custom data (\ref{sec:ownData}), and c) sharing datasets and analysis functions (\ref{sec:ownData}).


\subsection{User model performance evaluation using preloaded datasets}\label{subsec:generic_analysis}

This section demonstrates how to use \lib{} for detailed analysis of the strengths and weaknesses of existing offensive language detection models.
For this demo, we use roBERTa-offensive \citep{DBLP:journals/corr/abs-2010-12421}, a pre-trained generic language model, fine-tuned on the SemEval2019 OffensEval dataset \cite{zampieri-etal-2019-semeval}.
We use this model to make top-level predictions (offensive or not) for both COLD and HateCheck.

Each individual error analysis function shows the model's performance with respect to a particular feature (i.e. an existing dataframe column, or a new one added by the function).
Each function returns two dataframes.
The \texttt{metrics} dataframe contains a classification report for the analysis.\footnote{Appendix \ref{sec:appendix} provides more code examples for loading in data and starting generic analyses.}
This dataframe uses \lib{}'s built in \texttt{Metrics} function, which is built upon and uses Scikit-learn's \cite{scikit-learn} metrics library.

The \texttt{plot\_info} dataframe contains details of the analysis for the selected dimension, plus computed accuracy and the option to show textual examples.
If \texttt{show\char`_examples = True}, the function returns one randomly-selected incorrectly-classified instance for each value of the dimension being analyzed.\footnote{The variable \texttt{show\char`_examples} defaults to false to avoid accidental viewing of hateful or offensive language.}
If the plot option is selected, the plots are built from the \texttt{plot\_info} dataframe.

\subsubsection{Generic analysis functions}
\input{tables/coarse_cold}
\input{tables/anno_agree_0}
Table~\ref{table:coarse_cold_tab} shows the classification report for roBERTa-offensive on COLD.
Here, the classification report provides F1, precision, and recall for the two categories of offensive and non-offensive, as well as the macros and weighted averages. 
This model performs better overall on offensive instances, with high recall, but shows much better precision for non-offensive instances.
These reports can be easily modified to analyze subsets of the data.

Table \ref{table:anno_agree_tab} shows \texttt{plot\char`_info} for roBERTa-offensive on COLD, using 
\texttt{check\_anno\_agreement}.
The table shows accuracy for each category (full vs. partial) and one example incorrect prediction.
Accuracy is much higher for instances with full agreement than for those with some disagreement.
Offensiveness can be subjective, so it is useful to examine model performance on these different cases. 
Showing examples allows users to review difficult cases and may provide insights for model improvement.

\input{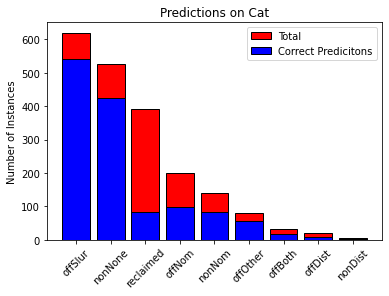}

\subsubsection{COLD analysis}

\input{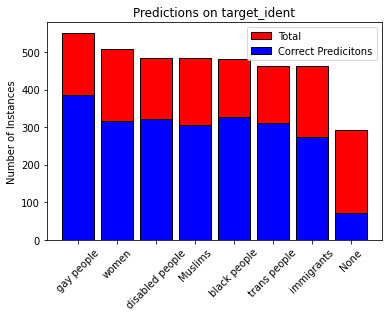} 

\input{imgs/Plots/compare_models_on_hc}
\noindent
The next analysis (Fig. \ref{fig:catCOLDplot}) is on the fine-grained COLD categories. 
roBERTa performs well on offensive tweets containing slurs and poorly on the reclaimed category (non-offensive tweets containing slurs).
The model also performs poorly on offensive tweets containing adjectival nominalization. 
These insights suggest that the model relies too much on slurs for identifying offensive language.

\subsubsection{HateCheck analysis}

HateCheck includes a column identifying the targeted group of the offensive language. 
This analysis can help identify which groups a model is more or less sensitive to. 
Figure \ref{fig:target_ident_HCplot} shows that the model does a better job of recognizing hateful speech when an identity is targeted than when one is not.



Figure \ref{subfig:roberta_hatecheck} analyzes the performance of roBERTa-offensive on the fine-grained HateCheck categories; complete table is in Appendix~\ref{app:metrics}. 
Most instances of counter-speech (when hate speech is quoted or is referred to with a non-hateful effect) are incorrectly classified.
An example of counter-speech is "I lost all respect for you when you said women are pathetic". 
Instances belonging to the categories of derogation, identity, threats, pronoun references, and phrasing are mostly correctly classified.

\subsubsection{Model comparison}\label{sec:compare}
So far we have compared one model's performance on two different datasets. 
Next, we compare two different models on the same dataset.
The second model is HateXplain \cite{mathew2021hatexplain}.
Both models of course suffer when classifying data sourced from outside their training domain.\footnote{roBERTa-offensive reports an F1 of 0.78 on OLID but drops to 0.62 on HateCheck. HateXplain reports F1 of 0.69 on the HateXplain dataset, and drops to 0.37 on HateCheck.}\textsuperscript{,}\footnote{We map HateXplain's ``offensive'' and ``hate speech'' labels both to HateCheck's ``hateful''.}
But we get a much truer sense of the models' out-of-domain capabilities when we look at the side-by-side analysis of the two models for the fine-grained HateCheck categories (Figures \ref{subfig:roberta_hatecheck} and \ref{subfig:hatexplain_hatecheck}).
While roBERTa-offensive does not perform well on counter-speech, HateXplain correctly classifies most counter-speech instances. And while HateXplain struggles to recognize hateful expressions with spelling changes, roBERTa does much better. 

\subsection{\lib{} as infrastructure: Extending functionality}\label{sec:ownData}

\lib{} is open-source\footnote{\url{https://github.com/alexispalmer/olea}, Licensed under MIT License} and has been designed to be extensible with new datasets and new analyses.

\subsubsection{Analysis on custom data}



The analysis methods described above can be easily applied to new corpora.
The code below shows the process of loading the OLID dataset \cite{zampierietal2019} as a pandas dataframe.
The user only needs to specify a path to the data and the relevant column headings.
The \texttt{Dataset} class acts as a wrapper for the data loaded from disk and allows the user to access class utilities such as \texttt{generator()}, which in turn is helpful for accessing data in batches.

\input{code_examples/section_4_code_examples/extend_olid}


\noindent
We can now submit model predictions, returning a \texttt{DatasetSubmissionObject} which can be used to conduct the generic analyses previously described; code in Appendix~\ref{sec:appendix}.


\subsubsection{Sharing datasets and analysis patterns} \label{sec:sharing}

With just a bit of coding, interfaces new datasets can be added to the \lib{} library more permanently, and for the benefit of all users.\footnote{\lib{} is not currently hosting datasets. The preloaded datasets are hosted via HuggingFace's datasets library.}
We demonstrate again using OLID,\footnote{Note that we only consider OLID's "level-A" annotations.} establishing the new \texttt{OLIDDataset} class which inherits from \texttt{Dataset}.

\input{code_examples/section_4_code_examples/olid_class}

\noindent
To accommodate the properties of the new dataset, we need to override some attributes of the \texttt{Dataset} class and to modify the method for loading data.

\lib{}'s scaffolding minimizes the amount of new code needed to add a new dataset, as well as automatically handling helper utilities such as mapping model predictions to the custom \texttt{Dataset} object.
The library also runs sanity checks on submitted predictions before returning a \texttt{DatasetSubmissionObject}.

The advantage of using a native \texttt{DatasetSubmissionObject} is that users may run \texttt{Generic} analyses on it. 
However, if authors have a unique analysis that they wish to couple with their dataset, they may specify a special \texttt{Analysis} class that can operate on submissions. 
The class methods can be modified to accommodate patterns or properties specific to the dataset.

\input{code_examples/section_4_code_examples/olid_analysis}

%



%% file: tables/coarse_cold.tex
\begin{table}[t!]\captionsetup{justification=justified}
\resizebox{0.49\textwidth}{!}{\begin{tabular}{|l|l|l|l|l|}
\hline
          & N     & Y     & macro avg & weighted avg \\ \hline
precision & 0.743 & 0.587 & 0.665     & 0.670        \\ \hline
recall    & 0.502 & 0.803 & 0.652     & 0.643        \\ \hline
f1-score  & 0.599 & 0.678 & 0.639     & 0.636        \\ \hline
support   & 1072  & 944   & 2016      & 2016         \\ \hline
\end{tabular}}
\caption{\texttt{Metrics} classification report for roBERTa-offensive on COLD, using \texttt{analyze\_on} on the dimension of offensiveness. N$=$not offensive, Y$=$offensive.}
\label{table:coarse_cold_tab}
\end{table}

%% file: tables/anno_agree_0.tex
\begin{table}[!t]\centering\captionsetup{justification=justified}
\resizebox{0.49\textwidth}{!}{\begin{tabular}{|c|c|c|}
\hline
\textbf{Annotator Agreement}                                                & \textbf{Full}                                                                                                                                                                     & \textbf{Partial}                                                                                                                                                                    \\ \hline
Total                                                                       & 1431                                                                                                                                                                              & 585                                                                                                                                                                                 \\ \hline
Total Correct                                                               & 1004                                                                                                                                                                              & 292                                                                                                                                                                                 \\ \hline
Accuracy                                                                    & 0.702                                                                                                                                                                             & 0.499                                                                                                                                                                               \\ \hline
\begin{tabular}[c]{@{}c@{}}Incorrect Classification \\ Example\end{tabular} & \multicolumn{1}{l|}{\begin{tabular}[c]{@{}l@{}}an illegal is not an \\ immigrant and \\ illegals do take \\ american jobs \\ considering they \\ are not americans.\end{tabular}} & \multicolumn{1}{l|}{\begin{tabular}[c]{@{}l@{}}USER yooo i was \\ thinking bout \\ that the other day \\ lol.. you only really \\ speaking of one \\ person my n*ggah\end{tabular}} \\ \hline
\begin{tabular}[c]{@{}c@{}}Example's \\ Predicted Label\end{tabular}        & N                                                                                                                                                                                 & Y                                                                                                                                                                                   \\ \hline
Example's Gold Label                                                        & Y                                                                                                                                                                                 & N                                                                                                                                                                                   \\ \hline
\end{tabular}}
\caption{\texttt{plot\_info} report for roBERTa-offensive on COLD, using \texttt{check\_anno\_agreement} (full vs. partial), with randomly-selected examples.}
\label{table:anno_agree_tab}
\end{table}

%% file: imgs/Plots/cat-roberta-cold.tex
\begin{figure}[th]\captionsetup{justification=justified}
\includegraphics[width=0.5\textwidth]{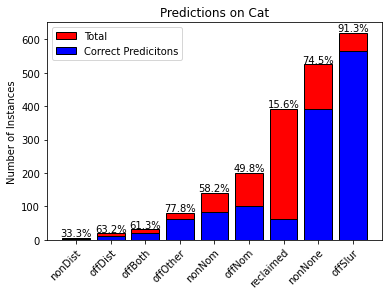}
\caption{Results for roBERTa-offensive on COLD, focusing on fine-grained categories. Percent value above a bar shows percent accuracy for that category.}
\centering
 \label{fig:catCOLDplot}
\end{figure}

%% file: imgs/Plots/target_ident_hc.tex
\begin{figure}[t]\captionsetup{justification=justified}
\includegraphics[width=0.5\textwidth]{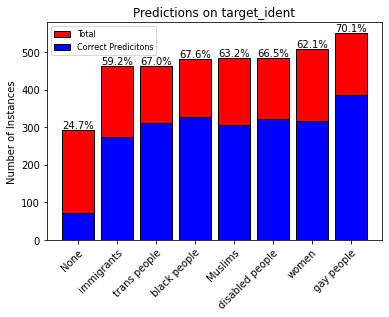}
\caption{Results for roBERTa-offensive on HateCheck, focusing on identity of the target.}
\centering
 \label{fig:target_ident_HCplot}
\end{figure}

%% file: imgs/Plots/compare_models_on_hc.tex
\begin{figure*}[ht!]\captionsetup{justification=justified}
  \caption{Comparison of HateXplain and roBERTa models on the fine-grained HateCheck categories. The category labels are followed by either (h) to denote that it is comprised of only hateful instances, (nh) for only non-hateful instances, or nothing to denote a mix of hateful and non-hateful instances}
  \centering
  \begin{subfigure}[t]{0.49\linewidth}
    \includegraphics[width=\linewidth]{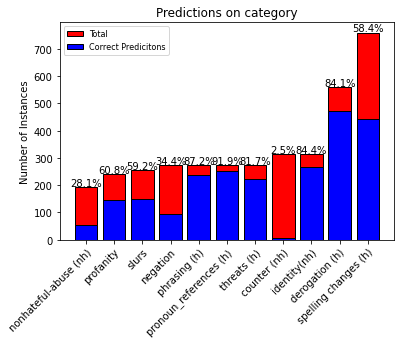}
    \caption{Results for roBERTa-offensive on HateCheck, focusing on fine-grained categories.}
    \label{subfig:roberta_hatecheck}
  \end{subfigure}
  \begin{subfigure}[t]{0.49\linewidth}
    \includegraphics[width=\linewidth]{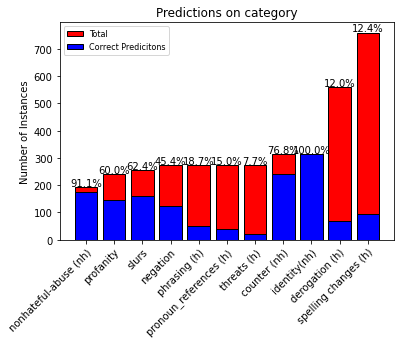}
    \caption{Results for HateXplain on HateCheck, focusing on fine-grained categories.}
    \label{subfig:hatexplain_hatecheck}
  \end{subfigure}
  \label{fig:compare_models}
\end{figure*}

%% file: code_examples/section_4_code_examples/extend_olid.tex
\begin{lstlisting}[language=Python]
olid = pd.read_csv('data/olid/olid_levela.csv')
olid_dataset = Dataset(data = olid, 
                features = 'Text', 
                gold_column = 'label', 
                text_column = 'Text')
data_gen = olid_dataset.generator(batch_size=64)
data = next(data_gen)
\end{lstlisting}

%% file: code_examples/section_4_code_examples/olid_class.tex
\begin{lstlisting}[language=Python]
class OLIDDataset(Dataset) : 
    text_column = 'Text'
    gold_column = 'label'
    features = ['Text','label_id']
def __init__(self, olid_csv_path:str) :
    self.olid_csv_path = olid_csv_path
    self._data = self._load_data()
def _load_data(self) -> pd.DataFrame:
    return pd.read_csv(self.olid_csv_path)
\end{lstlisting}

%% file: code_examples/section_4_code_examples/olid_analysis.tex
\begin{lstlisting}[language=Python]
class OLIDAnalysis(object) : 
    @classmethod
    def analyze_on(cls, submission:DatasetSubmissionObject, on:str) : 
        '''
        Unique OLID analysis goes here!
        '''
        return get_metrics(submission, on)
\end{lstlisting}

%% file: tex/discussion_conclusion.tex
\section{Conclusion and Future Directions}\label{sec:disc_conc}

This paper introduces \lib{}, a tool for easy, in-depth error analysis and an infrastructure for sharing new datasets and analysis methods.
\lib{} helps researchers understand the strengths and weaknesses of their offensive language detection models.
In the near term, we will continue to add new analysis methods and datasets, including methods for corpus exploration.
Mid term, we plan to extend \lib{} to additional languages, and eventually we would like to expand OLEA into a general error analysis library for a range of language classification tasks.
Because \lib{} is a convenient way for authors to share datasets and analyses, it is our hope that a community will develop around the library, and that models ultimately will improve as we learn more about what they can and cannot do.

%% file: tex/endmatter.tex
\section*{Limitations}






\lib{} is restricted in the languages that it can be used on. 
Currently, it assumes for some of the analyses on substrings that the language is delimited by spaces.  
Additionally, the library is primarily focused on providing error analyses for offensive language applications.  
Its use outside of this domain is not known or well defined.  
Though we focus on error analysis for offensive language identification, the system makes no binding assumptions as to the proper definitions of offensive language and hate speech, nor does it assert (or assume) any difference between these two categories. 
This lack of distinction complicates cross-model comparison.
Furthermore, this tool is just an analysis tool; it does not address concerns regarding language drift and other sociolinguistic biases that may be present within a user's dataset, nor does it address any annotator biases present in original datasets. 
OLEA is not a fully independent package as it relies on the external site HuggingFace for dataset hosting.





\section*{Ethics Statement}
This work provides important insights on hate speech and offensive language models for users, but we acknowledge the ethical implications of releasing a tool that encourages accessing hate speech datasets where tweet author anonymity may not be ensured. We attempt to take steps to minimize user exposure to hateful or offensive language when possible, by defaulting \texttt{show\char`_examples} to \texttt{False} during analysis. This tool is intended to help researchers understand their model performance, and should not be used for any task that promotes or spreads the usage of hate speech or unnecessarily exposes people to hate speech.


%% file: tex/appendix.tex
\appendix
\section{Code Examples}
\label{sec:appendix}
In this appendix, we provide code snippets corresponding to Section \ref{sec:usecases}.

\subsection{Preliminaries}
Import Statements
\input{code_examples/section_4_code_examples/import_statements}

\subsection{Code examples for Section~\ref{subsec:generic_analysis}}
\noindent Download and Preprocess COLD text:
\input{code_examples/section_4_code_examples/roberta_preprocess}
Create predictions using roBERTa-offensive
\input{code_examples/section_4_code_examples/roberta_load_predict}
Create Submission Objects:
\input{code_examples/section_4_code_examples/cold_submission}
\input{code_examples/section_4_code_examples/hc_submission}
Generate Table \ref{table:coarse_cold_tab}:
\input{code_examples/section_4_code_examples/cold_coarse_code}
Generate Table \ref{table:coarse_cold_tab} and Save plot to file:
\input{code_examples/section_4_code_examples/table_1_and_save_fig}
Generate Table \ref{table:anno_agree_tab}:
\input{code_examples/section_4_code_examples/cold_anno_agree_code}
Generate Figure \ref{fig:catCOLDplot}:
\input{code_examples/section_4_code_examples/cold_cat_code}
Generate Figure \ref{fig:target_ident_HCplot}:
\input{code_examples/section_4_code_examples/hc_target_ident_code}
Generate Figure \ref{subfig:roberta_hatecheck}, Figure \ref{subfig:hatexplain_hatecheck}, and Table \ref{HC_Cat_Metrics}:
\input{code_examples/section_4_code_examples/hc_cat_code}

\subsection{Code examples for Section~\ref{sec:ownData}}
Run analysis functions on local custom data:
\input{code_examples/section_4_code_examples/olid_predict}
Submit predictions for newly-established dataset class \texttt{OLIDDataset}:
\input{code_examples/section_4_code_examples/olid_submit}

\vfill\break
\section{Full Results Table for Figure~\ref{subfig:roberta_hatecheck}}\label{app:metrics}
\input{tables/hc_cat_1}

\vfill\break
\section{COLD: Features available for analysis}\label{app:coldDim}
\input{tables/COLD_Features.tex}

\section{HateCheck: Features available for analysis}\label{app:hcDim}
\input{tables/hc_features}

%% file: code_examples/section_4_code_examples/import_statements.tex
\begin{lstlisting}[language=Python]
import pandas as pd
from olea.data.cold import COLD, COLDSubmissionObject
from olea.data.hatecheck import HateCheck
from olea.analysis.cold import COLDAnalysis
from olea.analysis.generic import Generic
from olea.analysis.hatecheck import HateCheckAnalysis
from olea.utils import preprocess_text

\end{lstlisting}

%% file: code_examples/section_4_code_examples/roberta_preprocess.tex
\begin{lstlisting}[language=Python]
cold = COLD()
pt = preprocess_text.PreprocessText()
processed_text = pt.execute(cold.data()["Text"])
cold.data()["Text"] =processed_text
\end{lstlisting}

%% file: code_examples/section_4_code_examples/roberta_load_predict.tex
 \begin{lstlisting}[language=Python]
from transformers import AutoTokenizer, AutoModelForSequenceClassification
from transformers import TextClassificationPipeline
 
 link = "cardiffnlp/twitter-roberta-base-offensive"
 tokenizer = AutoTokenizer.from_pretrained(link)
        model = AutoModelForSequenceClassification.from_pretrained(link)
        #Create Pipeline for Predicting
        pipe = TextClassificationPipeline(model=model, tokenizer=tokenizer)
        preds = pd.DataFrame(pipe(list(cold.data()["Text"])))["label"]
\end{lstlisting}

%% file: code_examples/section_4_code_examples/cold_submission.tex
\begin{lstlisting}[language=Python]
cold_so = cold.submit(
    batch = cold.data(),
    predictions = preds,
    map = {"LABEL_0": 'N', 'LABEL_1': "Y"})
\end{lstlisting}

%% file: code_examples/section_4_code_examples/hc_submission.tex
\begin{lstlisting}[language=Python]
hc_so = hc.submit(
    batch = hc.data(),
    predictions = preds,
    map = {"LABEL_0": 'non-hateful', 'LABEL_1': "hateful"})
\end{lstlisting}

%% file: code_examples/section_4_code_examples/cold_coarse_code.tex
\begin{lstlisting}[language=Python]
plot_info, metrics = Generic.analyze_on(
    cold_so,
    'Cat',
    show_examples = False,
    plot = False)
\end{lstlisting}

%% file: code_examples/section_4_code_examples/table_1_and_save_fig.tex
\begin{lstlisting}[language=Python]
plot_info, metrics = Generic.analyze_on(
    cold_so,
    'Cat',
    show_examples = False,
    plot = False,
    savePlotToFile= "cold_cats.png")
\end{lstlisting}

%% file: code_examples/section_4_code_examples/cold_anno_agree_code.tex
\begin{lstlisting}[language=Python]
plot_info, metrics = Generic.check_anno_agreement(cold_so, ["Off1","Off2","Off3"],show_examples = True, plot = False)
\end{lstlisting}

%% file: code_examples/section_4_code_examples/cold_cat_code.tex
\begin{lstlisting}[language=Python]
plot_info, metrics =COLDAnalysis.analyze_on(
    cold_so,
    'Cat',
    show_examples = False, 
    plot = True)
\end{lstlisting}

%% file: code_examples/section_4_code_examples/hc_target_ident_code.tex
\begin{lstlisting}[language=Python]
plot_info, metrics = Generic.analyze_on(
    hc_so,
    'target_ident')
\end{lstlisting}

%% file: code_examples/section_4_code_examples/hc_cat_code.tex
\begin{lstlisting}[language=Python]
plot_info, metrics = HateCheckAnalysis.analyze_on(
    hc_so,
    'category')
\end{lstlisting}

%% file: code_examples/section_4_code_examples/olid_predict.tex
\begin{lstlisting}[language=Python]
predictions = model.predict(data)
submission = olid_dataset.submit(
            batch = data, 
            predictions = predictions, 
            map = {1:'OFF', 0:'NOT'})
# performance on AAVE
Generic.aave(submission) 
# performance on texts containing substring 'female'
Generic.check_substring(submission, "female") 
\end{lstlisting}

%% file: code_examples/section_4_code_examples/olid_submit.tex
\begin{lstlisting}[language=Python]
olid = OLIDDataset('data/olid.csv')
datagen = olid.generator(64)
data = next(datagen)
preds = model.predict(data)
map = {'OFF' : 1.0 , 'NOT' : 0.0}
submission = olid.submit(batch = data, 
               predictions = preds, 
               map = map)
\end{lstlisting}

%% file: tables/hc_cat_1.tex
\begin{table}[!h]\centering\captionsetup{justification=justified}
\resizebox{\columnwidth}{!}{%
\begin{tabular}{|l|l|l|l|l|l|}
\hline
category & Metrics & precision & recall & f1-score & support \\ \hline
\rowcolor[HTML]{EFEFEF} 
counter (nh) & hateful & 0.000 & 0.000 & 0.000 & 0 \\ \hline
\rowcolor[HTML]{EFEFEF} 
counter (nh) & non-hateful & 1.000 & 0.038 & 0.074 & 314 \\ \hline
\rowcolor[HTML]{EFEFEF} 
counter (nh) & macro avg & 0.500 & 0.019 & 0.037 & 314 \\ \hline
\rowcolor[HTML]{EFEFEF} 
counter (nh) & weighted avg & 1.000 & 0.038 & 0.074 & 314 \\ \hline
derogation (h) & hateful & 1.000 & 0.805 & 0.892 & 560 \\ \hline
derogation (h) & non-hateful & 0.000 & 0.000 & 0.000 & 0 \\ \hline
derogation (h) & macro avg & 0.500 & 0.403 & 0.446 & 560 \\ \hline
derogation (h) & weighted avg & 1.000 & 0.805 & 0.892 & 560 \\ \hline
\rowcolor[HTML]{EFEFEF} 
identity(nh) & hateful & 0.000 & 0.000 & 0.000 & 0 \\ \hline
\rowcolor[HTML]{EFEFEF} 
identity(nh) & non-hateful & 1.000 & 0.892 & 0.943 & 315 \\ \hline
\rowcolor[HTML]{EFEFEF} 
identity(nh) & macro avg & 0.500 & 0.446 & 0.471 & 315 \\ \hline
\rowcolor[HTML]{EFEFEF} 
identity(nh) & weighted avg & 1.000 & 0.892 & 0.943 & 315 \\ \hline
negation & hateful & 0.295 & 0.236 & 0.262 & 140 \\ \hline
negation & non-hateful & 0.335 & 0.406 & 0.367 & 133 \\ \hline
negation & macro avg & 0.315 & 0.321 & 0.315 & 273 \\ \hline
negation & weighted avg & 0.315 & 0.319 & 0.313 & 273 \\ \hline
\rowcolor[HTML]{EFEFEF} 
nonhateful-abuse (nh) & hateful & 0.000 & 0.000 & 0.000 & 0 \\ \hline
\rowcolor[HTML]{EFEFEF} 
nonhateful-abuse (nh) & non-hateful & 1.000 & 0.339 & 0.506 & 192 \\ \hline
\rowcolor[HTML]{EFEFEF} 
nonhateful-abuse (nh) & macro avg & 0.500 & 0.169 & 0.253 & 192 \\ \hline
\rowcolor[HTML]{EFEFEF} 
nonhateful-abuse (nh) & weighted avg & 1.000 & 0.339 & 0.506 & 192 \\ \hline
phrasing (h) & hateful & 1.000 & 0.868 & 0.929 & 273 \\ \hline
phrasing (h) & non-hateful & 0.000 & 0.000 & 0.000 & 0 \\ \hline
phrasing (h) & macro avg & 0.500 & 0.434 & 0.465 & 273 \\ \hline
phrasing (h) & weighted avg & 1.000 & 0.868 & 0.929 & 273 \\ \hline
\rowcolor[HTML]{EFEFEF} 
profanity & hateful & 0.601 & 1.000 & 0.751 & 140 \\ \hline
\rowcolor[HTML]{EFEFEF} 
profanity & non-hateful & 1.000 & 0.070 & 0.131 & 100 \\ \hline
\rowcolor[HTML]{EFEFEF} 
profanity & macro avg & 0.800 & 0.535 & 0.441 & 240 \\ \hline
\rowcolor[HTML]{EFEFEF} 
profanity & weighted avg & 0.767 & 0.613 & 0.492 & 240 \\ \hline
pronoun-references (h) & hateful & 1.000 & 0.908 & 0.952 & 273 \\ \hline
pronoun-references (h) & non-hateful & 0.000 & 0.000 & 0.000 & 0 \\ \hline
pronoun-references (h) & macro avg & 0.500 & 0.454 & 0.476 & 273 \\ \hline
pronoun-references (h) & weighted avg & 1.000 & 0.908 & 0.952 & 273 \\ \hline
\rowcolor[HTML]{EFEFEF} 
slurs & hateful & 0.593 & 0.778 & 0.673 & 144 \\ \hline
\rowcolor[HTML]{EFEFEF} 
slurs & non-hateful & 0.515 & 0.306 & 0.384 & 111 \\ \hline
\rowcolor[HTML]{EFEFEF} 
slurs & macro avg & 0.554 & 0.542 & 0.528 & 255 \\ \hline
\rowcolor[HTML]{EFEFEF} 
slurs & weighted avg & 0.559 & 0.573 & 0.547 & 255 \\ \hline
spelling changes (h) & hateful & 1.000 & 0.549 & 0.709 & 760 \\ \hline
spelling changes (h) & non-hateful & 0.000 & 0.000 & 0.000 & 0 \\ \hline
spelling changes (h) & macro avg & 0.500 & 0.274 & 0.354 & 760 \\ \hline
spelling changes (h) & weighted avg & 1.000 & 0.549 & 0.709 & 760 \\ \hline
\rowcolor[HTML]{EFEFEF} 
threats (h) & hateful & 1.000 & 0.810 & 0.895 & 273 \\ \hline
\rowcolor[HTML]{EFEFEF} 
threats (h) & non-hateful & 0.000 & 0.000 & 0.000 & 0 \\ \hline
\rowcolor[HTML]{EFEFEF} 
threats (h) & macro avg & 0.500 & 0.405 & 0.447 & 273 \\ \hline
\rowcolor[HTML]{EFEFEF} 
threats (h) & weighted avg & 1.000 & 0.810 & 0.895 & 273 \\ \hline
\end{tabular}%
}
\caption{The \texttt{Metrics} classification report for roBERTa-offensive on HateCheck}
\label{HC_Cat_Metrics}
\end{table}

%% file: tables/COLD_Features.tex
\begin{table}[h!]
\resizebox{0.49\textwidth}{!}{\begin{tabular}{|c|c|}
\hline
\rowcolor[HTML]{EFEFEF} 
\textbf{Feature}    & \textbf{Description}                                                                                                      \\ \hline
ID                  & The unique ID for the text                                                                                                \\ \hline
\rowcolor[HTML]{EFEFEF} 
Text                & \begin{tabular}[c]{@{}c@{}}The text containing social media messages \\ (some containing offensive language)\end{tabular} \\ \hline
Cat                 & The gold label category of the text                                                                                       \\ \hline
\rowcolor[HTML]{EFEFEF} 
Off                 & \begin{tabular}[c]{@{}c@{}}Offensive or not? ( Y / N )\\ Majority Vote\end{tabular}                                       \\ \hline
\rowcolor[HTML]{EFEFEF} 
Off1, Off2, Off3    & Individual annotator labels for Off ( Y / N )                                                                             \\ \hline
Slur                & \begin{tabular}[c]{@{}c@{}}Contains a slur?  ( Y / N )\\ Majority Vote\end{tabular}                                       \\ \hline
Slur1, Slur2, Slur3 & Individual annotator labels for Slur ( Y / N )                                                                            \\ \hline
\rowcolor[HTML]{EFEFEF} 
Nom                 & \begin{tabular}[c]{@{}c@{}}Contains adjectival nominalization? ( Y / N )\\ Majority Vote\end{tabular}                     \\ \hline
\rowcolor[HTML]{EFEFEF} 
Nom1, Nom2, Nom3    & Individual annotator labels for Slur ( Y / N )                                                                            \\ \hline
Dist                & \begin{tabular}[c]{@{}c@{}}Contains linguistic distancing?  ( Y / N )\\ Majority Vote\end{tabular}                        \\ \hline
Dist1, Dist2, Dist3 & Individual annotator labels for Dist ( Y / N )                                                                            \\ \hline
\end{tabular}}
\caption{Features available as part of the COLD dataset. 
The feature names and descriptions come from \citet{palmer2020cold}.}
\label{coldFeatures}
\end{table}

%% file: tables/hc_features.tex
\begin{table}[h!]
\resizebox{0.49\textwidth}{!}{%
\begin{tabular}{|c|l|}
\hline
\textbf{Feature}                  & \multicolumn{1}{c|}{\textbf{Description}}                                                                                                                                                                                                                                     \\ \hline
functionality                     & \begin{tabular}[c]{@{}l@{}}The shorthand for the functionality tested \\ by the test case.\end{tabular}                                                                                                                                                                       \\ \hline
case\_id                          & \begin{tabular}[c]{@{}l@{}}The unique ID of the test case (assigned to \\ each of the 3,901 cases initially generated)\end{tabular}                                                                                                                                           \\ \hline
test\_case                        & The text of the test case.                                                                                                                                                                                                                                                    \\ \hline
\multicolumn{1}{|l|}{label\_gold} & \begin{tabular}[c]{@{}l@{}}The gold standard label (hateful/non-hateful) \\ of the test case. All test cases within a given \\ functionality have the same gold standard label.\end{tabular}                                                                                  \\ \hline
target\_ident                     & \begin{tabular}[c]{@{}l@{}}Where applicable, the protected group \\ targeted or referenced by the test case. We \\ cover seven protected groups in the test suite: \\ women, trans people, gay people, black people, \\ disabled people, Muslims and immigrants.\end{tabular} \\ \hline
direction                         & \begin{tabular}[c]{@{}l@{}}For hateful cases, the binary secondary label\\ indicating whether they are directed at an \\ individual as part of a protected group or aimed\\ at the group in general.\end{tabular}                                                             \\ \hline
focus\_words                      & \begin{tabular}[c]{@{}l@{}}Where applicable, the key word or phrase in a \\ given test case (e.g. "cut their throats").\end{tabular}                                                                                                                                          \\ \hline
focus\_lemma                      & \begin{tabular}[c]{@{}l@{}}Where applicable, the corresponding \\ lemma (e.g. "cut sb. throat").\end{tabular}                                                                                                                                                                 \\ \hline
ref\_case\_id                     & \begin{tabular}[c]{@{}l@{}}For hateful cases, where applicable, the ID of the \\ simpler hateful case which was perturbed to \\ generate them. For non-hateful cases, where \\ applicable, the ID of the hateful case which is \\ contrasted.\end{tabular}                    \\ \hline
ref\_templ\_id                    & The equivalent, but for template IDs.                                                                                                                                                                                                                                         \\ \hline
templ\_id                         & \begin{tabular}[c]{@{}l@{}}The unique ID of the template from which the \\ test case was generated (assigned to each of the \\ 866 cases and templates from which we \\ generated the 3,901 initial cases).\end{tabular}                                                      \\ \hline
\end{tabular}
}
\caption{Features available as part of the HateCheck dataset. Feature names and descriptions come from \citet{rottger2020hatecheck}.}
\label{HC_features}
\end{table}